\documentclass[12pt, a4paper]{article}

\usepackage[margin=1in]{geometry}
\usepackage{amsmath, amssymb}
\usepackage{amsthm} 
\usepackage{graphicx}
\usepackage{xcolor}
\usepackage{tikz}
\usetikzlibrary{shapes.geometric, arrows.meta, positioning}
\usepackage{hyperref}
\usepackage{microtype} 
\usepackage{listings} 

\newtheorem{theorem}{Theorem}
\newtheorem{lemma}{Lemma}
\newtheorem{definition}{Definition}
\newtheorem{remark}{Remark}

\tikzset{
    block/.style={rectangle, draw=black, thick, fill=blue!5, text width=3.5cm, text centered, rounded corners, minimum height=1.5cm, align=center},
    subblock/.style={rectangle, draw=black, thick, fill=green!5, text width=3.5cm, text centered, rounded corners, minimum height=1.5cm, align=center},
    jsonblock/.style={rectangle, draw=black, thick, fill=gray!10, text width=3.5cm, text centered, rounded corners, minimum height=1.5cm, align=center},
    aiblock/.style={rectangle, draw=black, thick, fill=orange!10, text width=4cm, text centered, rounded corners, minimum height=1.5cm, align=center},
    engineblock/.style={rectangle, draw=black, very thick, fill=blue!15, text width=3.5cm, text centered, rounded corners, minimum height=1.5cm, align=center},
    bidir/.style={<->, >=Stealth, thick}
}

\title{Al-Ghasali: A Computational Framework for Simulating Islamic Jurisprudential Rulings via Finite State Automata and Large Language Models}

\author{Elnaserledinellah Mahmoud Abdelwahab\\
\href{mailto:elnaser@computational-fiqh.org}
{\texttt{elnaser@computational-fiqh.org}}}

\date{} 

\begin{document}

\maketitle

\begin{abstract}
This paper presents Al-Ghasali, a computational framework designed to formalize and simulate Islamic jurisprudential (Fiqh) rulings. Building upon recent set-theoretic formalizations of al-sabr wa al-Taqsim that extract minimal operational rules as simple Horn-clause logical programs, this work proves that the semantics of closed Fiqh chapters collapse completely to deterministic Finite State Automata (FSA). We establish a formal computational hierarchy, demonstrating that restricting formalization to finite propositional Horn theories guarantees decidability and constant-time (O(1)) verdict extraction after offline compilation. The framework formally proves that hierarchical compositions of sub-chapter FSAs preserve this O(1) complexity, and that temporal juristic requirements, such as Tartib (order), are deterministically flattened into static ontological states, preserving the memoryless property of the automaton. To generate human-readable juristic summaries, Al-Ghasali incorporates Large Language Models (LLMs) via an architecture where the deterministic FSA strictly controls the probabilistic LLM. Crucially, the paper formulates sufficient design safety conditions for this FSA-LLM integration, ensuring strict adherence to the epistemological boundaries of Usul al-Fiqh. To neutralize the risk of LLM hallucination, the system employs a Grammar-Guided Constrained Decoding mechanism. We formally define the mapping function that dynamically constructs unambiguous right-linear grammars from the FSA execution history. The paper mathematically proves that this architecture, augmented by a Bounded Neutral Lexicon and Valence Locking, guarantees the absolute absence of Decisional Hallucination, Juristic Factual Hallucination, and Semantic Inversion---reducing the LLM to a pure linguistic transformer of deterministic juristic facts.
\end{abstract}


\section{Introduction}
Recent advancements in computational Fiqh, particularly the set-theoretic formalization of the classical \textit{usuli} method of \textit{al-sabr wa al-Taqsim} (Examination and Division), have successfully introduced algorithms to extract minimal operational rules from complete juristic truth tables \cite{abdelwahab2026}. This framework operates under the Closed-World Assumption (CWA), bounding a Fiqh chapter to a finite Ontological Concepts Tree (OCT) of size $n$, which guarantees a finite legal state space of $2^n$. 

While the extracted rules provide a rigorous logical foundation, their direct execution in an interactive, user-facing application typically requires an embedded logical inference engine. In a dynamic environment where users simultaneously toggle multiple conditions, runtime logical resolution introduces unnecessary computational overhead and state-management complexity. 

To bridge the gap between theoretical logical extraction and practical deployment, we propose converting these extracted rules into Finite State Automata (FSA). We argue that FSA represent the best and most efficient computational model for this domain due to their deterministic $O(1)$ evaluation, natural modeling of non-monotonicity (preventers), and stateful mapping of sequential user interactions.

In this paper, we introduce \textit{Al-Ghasali}, a system that implements this FSA paradigm. It allows users to dynamically generate a UI matrix, link variables to embedded FSA rule sets, and evaluate final rulings based on algorithmic truth tables. Additionally, it utilizes LLMs to provide human-readable juristic summaries of the computational output, governed by strict architectural principles that mathematically eliminate the risk of LLM juristic hallucination.

The foundational premise of this framework relies on the ``Closed Chapter Assumption,'' which is not merely a mathematical convenience but an accurate reflection of the epistemological boundaries inherent to specific domains of Islamic jurisprudence. In particular, the chapters of Ibadat (ritual worship)---such as purification (Tahara), prayer (Salah), fasting (Sawm), and pilgrimage (Hajj)---are strictly Tawqifi \cite{alshatibi1997, kamali2003}, meaning their legal ontologies are divinely bounded by the Quran and Sunna. Consequently, independent juristic reasoning (ijtihad) operates strictly within this fixed conceptual vocabulary; no future jurist can invent a novel type of prayer or fasting, ensuring the Ontological Concepts Tree (OCT) for these chapters remains permanently stable. Even when unprecedented modern circumstances arise---such as space travel or synthetic materials---jurists do not expand the ontological state space, but rather map these new physical realities onto pre-existing, textually anchored legal categories. Therefore, because a significant portion of the Fiqh corpus operates within these permanently closed conceptual bounds, the state space is definitively constant, guaranteeing that the theoretical collapse of closed Horn chapters to a deterministic Finite State Automaton is not merely an abstract computational exercise, but a structurally sound and practically guaranteed model of juristic reality.

\section{Formalization}
Before describing the system architecture, we formally map the output of the \textit{al-sabr wa al-Taqsim} algorithm to strict Horn-clause logic, prove the adequacy of FSA for modeling these clauses, and define the sequential nature of Fiqh queries.

\subsection{Horn-Clause Representation of Usul Rules}
Let $V = \{v_1,\allowbreak v_2,\allowbreak \dots,\allowbreak v_n\}$ be the finite set of binary legally operative concepts (OCT) for a closed chapter, and let $H$ be the ruling atom. The algorithm in \cite{abdelwahab2026} extracts the set of positive minimal rules $M^+ = \{R_1, R_2, \dots, R_m\}$. The structural decomposition yields $R_i = S \land I_i$, where $S$ is the intersection of all $M^+$ (the \textit{Shurut}, conditions) and $I_i$ is the branch-specific component.

The algorithm defines the Global Candidate \textit{Illah} as the disjunction of all branch components: $I^* = I_1 \lor I_2 \lor \dots \lor I_m$. To express the ruling $H \leftarrow S \land I^*$ as a strict Horn program, we introduce an auxiliary atom $I$ representing the satisfaction of $I^*$, and we map the binary valuations of the OCT variables to distinct positive atoms (e.g., $v_i=1$ maps to atom $v_i^1$, and $v_i=0$ maps to atom $v_i^0$). The strict Horn-clause program $\mathcal{P}$ is constructed as follows:

\begin{align}
    I &\leftarrow I_i^+ \quad \forall I_i \in I^* \label{eq:illah} \\
    H &\leftarrow S^+ \land I \label{eq:ruling} \\
    &\leftarrow H \land W_j^+ \quad \forall W_j \in M^- \label{eq:mani}
\end{align}
where $I_i^+$, $S^+$, and $W_j^+$ are the conjunctions of positive atoms representing the respective binary valuations of the conditions, causes, and preventers. Equation~\ref{eq:illah} defines the auxiliary cause atom, Equation~\ref{eq:ruling} is the definite clause for the ruling, and Equation~\ref{eq:mani} represents the non-monotonic preventers (\textit{Mawani}) as Horn integrity constraints \cite{lloyd1987, alamidi2000, alshawkani1999}.

\subsection{Lemma: Finite-State Collapse of Closed Horn Chapters}

\begin{lemma}[Finite-State Collapse of Closed Horn Chapters] 
Let $\mathcal{P}$ be a strict Horn-clause program derived from a closed Fiqh chapter under the Closed-World Assumption (CWA). Assume:
\begin{itemize}
    \item The Ontological Concepts Tree (OCT) contains a finite set of binary variables $V=\{v_1,\allowbreak v_2,\allowbreak \dots,\allowbreak v_n\}$.
    \item Every clause of $\mathcal{P}$ is constructed solely from valuations of variables in $V$.
    \item $\mathcal{P}$ contains no function symbols, no existential variable generation, and no mechanism that expands the ontology beyond $V$.
\end{itemize}

Then:
\begin{enumerate}
    \item The semantic state space of $\mathcal{P}$ is finite and contains exactly $2^n$ possible valuations.
    \item Every execution of $\mathcal{P}$ is completely determined by the current valuation of $V$.
    \item There exists a deterministic finite-state automaton $\mathcal{A}=(Q, \Sigma, \delta, q_0, F)$ such that $Q=\{0,1\}^n$ and
    \[ \mathcal{P} \models H \iff q \in F. \]
    \item No computational model possessing memory beyond finite state (e.g. pushdown automata or Turing machines) provides additional expressive power for determining the ruling $H$.
\end{enumerate}
Hence the operational semantics of $\mathcal{P}$ collapse to finite-state behavior.

\textit{Proof.} 
Because the chapter is closed, the ontology consists of exactly the finite set $V=\{v_1, \dots, v_n\}$.

Each variable is binary, therefore every legal situation is represented by a valuation $s=(b_1, \dots, b_n) \in \{0,1\}^n$.

The number of distinct valuations is $|\{0,1\}^n| = 2^n$.
Hence the semantic universe of the chapter is finite.

By assumption, every clause in $\mathcal{P}$ refers only to variables in $V$. Since no function symbols or ontology-expanding operators exist, no new atoms can be generated during execution. Therefore every possible inference of $\mathcal{P}$ depends exclusively on the current valuation $s$.

Consequently, two executions reaching the same valuation $s$ are indistinguishable with respect to all future derivations of $\mathcal{P}$. The entire logical history may therefore be replaced by the current valuation without loss of information.

Define $Q=\{0,1\}^n$.
For each state $s \in Q$, evaluate the Horn program $\mathcal{P}$.
Let 
\[ F=\{s \in Q \mid \mathcal{P} \models H \text{ under valuation } s\}. \]
Since $Q$ is finite, $F$ is finite and can be precomputed.

Let state transitions correspond to changes in OCT variables. If an event changes valuation $s$ into valuation $s'$, define
\[ \delta(s, x) = s'. \]
This yields a deterministic finite-state automaton $\mathcal{A}=(Q, \Sigma, \delta, q_0, F)$.

For every valuation $s$,
\[ \mathcal{P} \models H \iff s \in F. \]
Thus the automaton exactly reproduces the semantics of the Horn program.

It remains to show that stronger computational models provide no additional expressive power.
A pushdown automaton exceeds finite-state power only through access to an unbounded stack. Likewise, a Turing machine exceeds finite-state power through access to unbounded memory. Under the present assumptions, however, every possible configuration of the chapter is already represented by one of the finitely many states in $Q$. There exists no operation capable of generating information outside this finite set.

Therefore any additional memory structure is observationally redundant: the truth of $H$ is a function solely of the current valuation
\[ f: \{0,1\}^n \to \{0,1\}. \]
Since finite-state machines can compute every function on a finite domain, the finite-state automaton already realizes the complete operational semantics of $\mathcal{P}$ \cite{hopcroft2006, sipser2013}.

Hence the Horn-clause program admits no computational behavior beyond finite-state behavior, and its semantics collapse to those of a deterministic finite-state automaton. $\square$ 
\end{lemma}

\subsection{Preservation of Decidability under Hierarchical Composition}
While Lemma 1 proves the finite-state collapse for a single closed chapter, the system architecture (Principle 1) allows a Main Chapter to utilize nested Sub-chapters (\textit{Bab Far'i}) to prevent combinatorial explosion. A Sub-chapter is an isolated FSA that evaluates a local ruling $H_{sub} \in \{0,1\}$ and passes this binary verdict back to the Main Chapter as a single input. We must prove that this hierarchical composition does not violate the finite-state collapse or the $O(1)$ extraction guarantee.\\

\begin{lemma}[Finite-State Preservation under Hierarchical Composition]
\label{lem:hierarchical}
Let $A_{main}$ be the deterministic FSA for a closed Main Chapter, characterized by its state, alphabet, and transition sets. Let $\{A_{sub}^{(i)}\}_{i=1}^k$ be a finite set of isolated Sub-chapter FSAs, each deriving a local verdict $H_{sub}^{(i)} \in \{0,1\}$. 

Assume:
\begin{itemize}
    \item The Main Chapter OCT contains a designated binary variable $v_{sub}^{(i)} \in V_{main}$ for each sub-chapter $i$.
    \item The valuation of $v_{sub}^{(i)}$ is strictly determined by the accepting state of $A_{sub}^{(i)}$, such that $v_{sub}^{(i)} = 1$ iff $H_{sub}^{(i)} = 1$.
    \item The composition is acyclic: Sub-chapters do not invoke the Main Chapter or create recursive loops.
\end{itemize}

Then the composite system $S$ preserves finite-state semantics, decidability, and constant-time ($O(1)$) verdict extraction for the Main Chapter ruling $H_{main}$.
\end{lemma}

\begin{proof}
By Lemma 1, each isolated Sub-chapter FSA $A_{sub}^{(i)}$ computes a deterministic boolean function over its finite state space. Because the sub-chapter's ontology size $n_i$ is finite, the computation of $H_{sub}^{(i)}$ is decidable and terminates in $O(1)$ time (as a precompiled lookup).

In the composite system $S$, the output of $A_{sub}^{(i)}$ is bound to the variable $v_{sub}^{(i)}$ in $V_{main}$. Because the composition is acyclic, there is no feedback loop from $A_{main}$ back into $A_{sub}^{(i)}$ during evaluation. Thus, $v_{sub}^{(i)}$ acts strictly as an atomic binary input.

Since $V_{main}$ is finite (size $n$), and every $v_{sub}^{(i)}$ resolves to a definitive binary value, the semantic state space of the Main Chapter remains exactly $Q_{main} = \{0,1\}^n$. The transition function $\delta_{main}$ treats the resolution of a sub-chapter as a state transition triggered by an input symbol $x$, updating $v_{sub}^{(i)}$ accordingly.

Because the Sub-chapter evaluation is bounded by $O(1)$, and the Main Chapter verdict extraction is bounded by $O(1)$, the total operational complexity to evaluate the Main Chapter, given all sub-chapters are resolved, remains $O(1)$. 

Furthermore, no new terms or atoms are generated outside the union of $V_{main}$ and $\bigcup V_{sub}^{(i)}$, so the Herbrand universe of the composite system remains finite. Thus, the hierarchical composition does not elevate the system to a Turing-complete or pushdown-automaton level; it strictly preserves the finite-state collapse and decidability guarantees of Lemma 1.
\end{proof}

\subsection{Formalization of Temporal Sequences and Juristic Order (\textit{Tartib})}
A critical theoretical distinction must be made regarding the temporal sequence of user inputs ($\sigma = \langle x_1, x_2, \dots, x_p \rangle$) and the evaluation of the ruling atom $H$. By definition, propositional Horn-clause logic is path-independent; the logical entailment $P \models H$ is evaluated strictly on the current semantic valuation $s \in \{0,1\}^n$, regardless of the order in which the variables reached that state. 

While many Fiqh chapters are purely state-dependent, certain rulings across various Madhhabs (schools of jurisprudence) impose \textit{Tartib} (juristic order) or \textit{Muwalah} (continuity). A canonical example is the requirement in the Shafi'i school that the acts of ablution (Wudu) be performed in a specific sequence \cite{alnawawi2002}, whereas the Hanafi school treats the sequence as non-binding \cite{almarghinani2004}. If the ruling $H$ were dependent on the sequence of toggles rather than the final valuation, the standard FSA state space $Q = \{0,1\}^n$ would be insufficient, as identical valuations reached via different sequences would yield contradictory rulings.

Because a deterministic Finite State Automaton is fundamentally memoryless, it cannot evaluate path-dependence without encoding that history into its current state. However, from the perspective of \textit{Usul al-Fiqh}, this computational necessity is perfectly aligned with juristic epistemology: in Madhhabs that require \textit{Tartib}, the sequence itself is legally classified as a condition (\textit{Shart}) or its absence as a preventer (\textit{Mani'}). Therefore, the sequence is not merely a physical property of the action, but a legally operative concept that natively belongs in the Ontological Concepts Tree (OCT).

To preserve the finite-state collapse and the $O(1)$ extraction guarantee, the framework formalizes juristic order by explicitly expanding the OCT to include a binary variable representing the satisfaction of the sequence, denoted as $v_{order} \in \{0,1\}$. Consequently, the state space $Q = \{0,1\}^n$ expands to include this dimension, but remains finite and strictly bounded.

The burden of tracking the temporal sequence is then shifted from the logical evaluation to the deterministic FSA transition function $\delta: Q \times \Sigma \rightarrow Q$. Let an input symbol $x \in \Sigma$ represent a user toggling a specific parameter. The transition function is constructed to enforce procedural jurisprudence by dynamically updating $v_{order}$.

To illustrate this, consider the Shafi'i requirement of washing the right arm ($v_{right}$) before the left arm ($v_{left}$) during \textit{Wudu}. Let $v_{order}$ be initialized to $1$. The state is represented as $s = (v_{right}, v_{left}, v_{order})$. The transition function $\delta$ operates as follows:

\begin{itemize}
    \item \textbf{Correct Sequence:} If the user toggles the right arm first from an initial state $s_0 = (0,0,1)$, the transition yields $\delta(s_0, \text{toggle\_right}) = (1,0,1)$. If the user subsequently toggles the left arm, the transition yields $\delta((1,0,1), \text{toggle\_left}) = (1,1,1)$. The final state $s_{final} = (1,1,1)$ maintains $v_{order} = 1$, satisfying the conditions of the Horn program $P$.
    \item \textbf{Incorrect Sequence:} If the user toggles the left arm first from the initial state $s_0$, the transition function detects that $v_{right} = 0$. It deterministically updates the state to $\delta(s_0, \text{toggle\_left}) = (0,1,0)$, setting $v_{order} = 0$. If the user subsequently toggles the right arm, the transition yields $\delta((0,1,0), \text{toggle\_right}) = (1,1,0)$. 
\end{itemize}

In the incorrect sequence, the final state $s_{final} = (1,1,0)$ contains the same primary action valuations ($v_{right}=1, v_{left}=1$) as the correct sequence, but differs critically in $v_{order}=0$. The extracted Horn program $P$ evaluates this static state, identifies $v_{order}=0$ as a preventer ($W$), and entails $H = 0$ (Invalid).

By this construction, the temporal sequence $\sigma$ is immediately "flattened" into a static boolean valuation upon each state transition. By the time the main parameter $b$ is toggled to evaluate $H$, the current state $s \in \{0,1\}^n$ already contains the boolean fact of whether the sequence was legally satisfied. Therefore, the path-dependence of \textit{Tartib} does not violate the memoryless property of the Horn-clause evaluation. The FSA deterministically maps temporal procedural violations into static ontological states, ensuring that Lemma 1 remains valid and that $O(1)$ verdict extraction is preserved even for strictly ordered juristic chapters.

\section{Decidability and Computational Complexity of Horn-Clause Fiqh Programs}

The computational efficiency of the proposed framework derives directly from the restricted form of the Horn-clause programs produced by the computational \textit{al-sabr wa al-Taqsim} algorithm. Unlike general logic programs, the extracted rules operate over a finite Ontological Concepts Tree (OCT) and contain neither recursive rule definitions nor function symbols. Consequently, they belong to the class of finite propositional Horn theories.

\subsection{Computational Properties of the Extracted Horn Programs}

Let
\[ V=\{v_1,\allowbreak v_2,\allowbreak \dots,\allowbreak v_n\} \]
be the finite set of binary legally operative concepts defining a closed Fiqh chapter. The corresponding Horn-clause program $\mathcal{P}$ consists exclusively of clauses whose atoms are drawn from $V$.

Because every variable is binary, every legal situation corresponds to a valuation
\[ s \in \{0,1\}^n. \]
The total number of semantic states is therefore finite and equal to
\[ |\{0,1\}^n| = 2^n. \]

Furthermore, since the program contains no function symbols and no recursive mechanisms capable of generating new concepts beyond those already present in the OCT, every possible inference depends solely upon the current valuation of the variables.

As shown in the finite-state-collapse lemma, the semantics of such programs reduce completely to finite-state behavior. The ruling atom $H$ can therefore be represented by the accepting-state set
\[ F = \{s \in \{0,1\}^n \mid \mathcal{P} \models H\}. \]

Since the state space is finite, the set $F$ may be computed offline during system initialization. Runtime verdict extraction then reduces to a simple membership test
\[ s \in F. \]

Consequently, after compilation, the asymptotic runtime complexity of ruling extraction becomes
\[ O(1). \]

The computational cost is shifted entirely to the offline compilation stage, while user interaction consists only of deterministic state transitions and constant-time verdict lookup.

This property provides one of the principal advantages of the framework. Because the chapter semantics collapse to a finite-state machine, no stronger computational model than a deterministic Finite State Automaton (FSA) is required for exact execution.

\subsection{Extension to Datalog-Style Recursive Rules}

Although the framework developed in this paper intentionally restricts itself to propositional Horn clauses, one may ask what happens if recursion is introduced while still forbidding function symbols. Such programs belong to the class of Datalog programs \cite{ceri1990}.

A natural jurisprudential example arises from the computation of genealogical relationships. Classical Fiqh contains numerous rulings that first determine a person's degree of kinship before applying the legal ruling itself. Examples include inheritance distribution, determination of agnatic relatives (\textit{\textquoteleft{}asabah}), maintenance obligations, guardianship, and the identification of prohibited degrees of marriage.

Consider a simplified genealogical database containing the predicate
\[ \text{parent}(x,y), \]
meaning that $x$ is a parent of $y$.

The ancestor relation may be defined recursively:
\[ \text{ancestor}(x,y) \leftarrow \text{parent}(x,y) \]
\[ \text{ancestor}(x,y) \leftarrow \text{parent}(x,z) \land \text{ancestor}(z,y). \]

Suppose that a particular inheritance rule applies to all descendants of a deceased person. The legal eligibility predicate may then be written as
\[ \text{eligible}(y) \leftarrow \text{ancestor}(x,y) \land \text{deceased}(x). \]

In this example, the legal ruling cannot be evaluated directly from a finite valuation of binary variables. Instead, the recursive ancestor relation must first be computed. The ruling is therefore obtained only after the recursive relation reaches its fixed point.

\subsection{Computational Consequences of Datalog Recursion}

The introduction of recursion significantly changes the computational model, but it does not destroy decidability.

Because Datalog still forbids function symbols, the Herbrand universe remains finite. No rule can generate new terms beyond those already present in the database. Consequently, every recursive derivation must eventually terminate at a least fixed point.

Therefore:
\begin{enumerate}
    \item Entailment remains decidable.
    \item Every recursive computation eventually terminates.
    \item Evaluation remains polynomial in the size of the underlying database.
\end{enumerate}

However, an important property is lost. The finite-state-collapse result no longer applies in its original form. The legal verdict can no longer be obtained solely through a precompiled lookup table over binary OCT valuations.

Instead, the system must first perform a fixed-point computation to derive all recursively implied relations before the final ruling can be evaluated.

The practical consequence is that verdict extraction is no longer a constant-time lookup operation. Rather, it requires execution of a recursive inference engine capable of computing the least model of the Datalog program.

\subsection{Beyond Datalog: Loss of Decidability}

A more dramatic change occurs when function symbols or unbounded term construction are introduced into Horn clauses.

For example, recursive constructions of the form
\[ P(x) \leftarrow P(f(x)) \]
allow the creation of infinitely many terms
\[ x, \quad f(x), \quad f(f(x)), \quad f(f(f(x))), \dots \]
producing an infinite Herbrand universe.

Such programs are capable of simulating pushdown automata and, in general, Turing machines. The finite-state semantics disappear completely, and entailment is no longer guaranteed to be decidable \cite{sipser2013}.

Thus the computational hierarchy relevant to Fiqh formalization may be summarized as
\[ \text{Propositional Horn Chapters} \subset \text{Datalog Chapters} \subset \text{General Recursive Horn Programs}. \]

The first level corresponds to the framework developed in this paper and enjoys finite-state semantics together with constant-time verdict extraction after compilation. The second level permits recursively defined juristic relations while preserving decidability and practical tractability. The third level introduces unbounded computational memory and generally sacrifices both finite-state behavior and decidability.

From a computational perspective, the present framework deliberately occupies the first level of this hierarchy. By restricting legal formalization to closed chapters with finite ontologies, the system retains the strongest possible computational guarantees: finite-state semantics, deterministic execution, decidability, and constant-time ruling extraction after compilation.

\section{System Architecture}
\begin{figure}[htbp]
    \centering
    \begin{tikzpicture}[node distance=2.5cm and 3cm]
        
        \node[engineblock] (engine) {\textbf{Evaluation Engine} \\ \small Rule Sync \& FSA Logic \\ \small Verdict Extraction};
        
        \node[block, above=of engine] (ui) {\textbf{Dynamic UI Layer} \\ \small $n \times m$ Matrix \\ \small Inputs \& Display};
        \node[jsonblock, left=of engine] (json) {\textbf{JSON Rule Sets} \\ \small Truth Tables \\ \small Hierarchical Hukm};
        \node[subblock, right=of engine] (sub) {\textbf{Sub-chapter Modal} \\ \small Isolated FSA \\ \small Context Locking};
        \node[aiblock, below=of engine] (ai) {\textbf{AI Integration Layer} \\ \small OpenAI-compatible API \\ \small Prompt Engineering};
        
        \draw[bidir] (ui) -- node[right, text width=1.5cm, align=center] {\scriptsize UI States \& Final Verdict} (engine);
        \draw[bidir] (engine) -- node[above, text width=1.5cm, align=center] {\scriptsize Recursive Sync} (json);
        \draw[bidir] (engine) -- node[above, text width=1.5cm, align=center] {\scriptsize Parent Context \& Sub-verdict} (sub);
        \draw[bidir] (engine) -- node[right, text width=1.5cm, align=center] {\scriptsize Prompt \& Summary} (ai);
        
    \end{tikzpicture}
    \caption{System Architecture of Al-Ghasali utilizing a Hub-and-Spoke model to demonstrate bidirectional data flow between the core engine and peripheral modules.}
    \label{fig:architecture}
\end{figure}
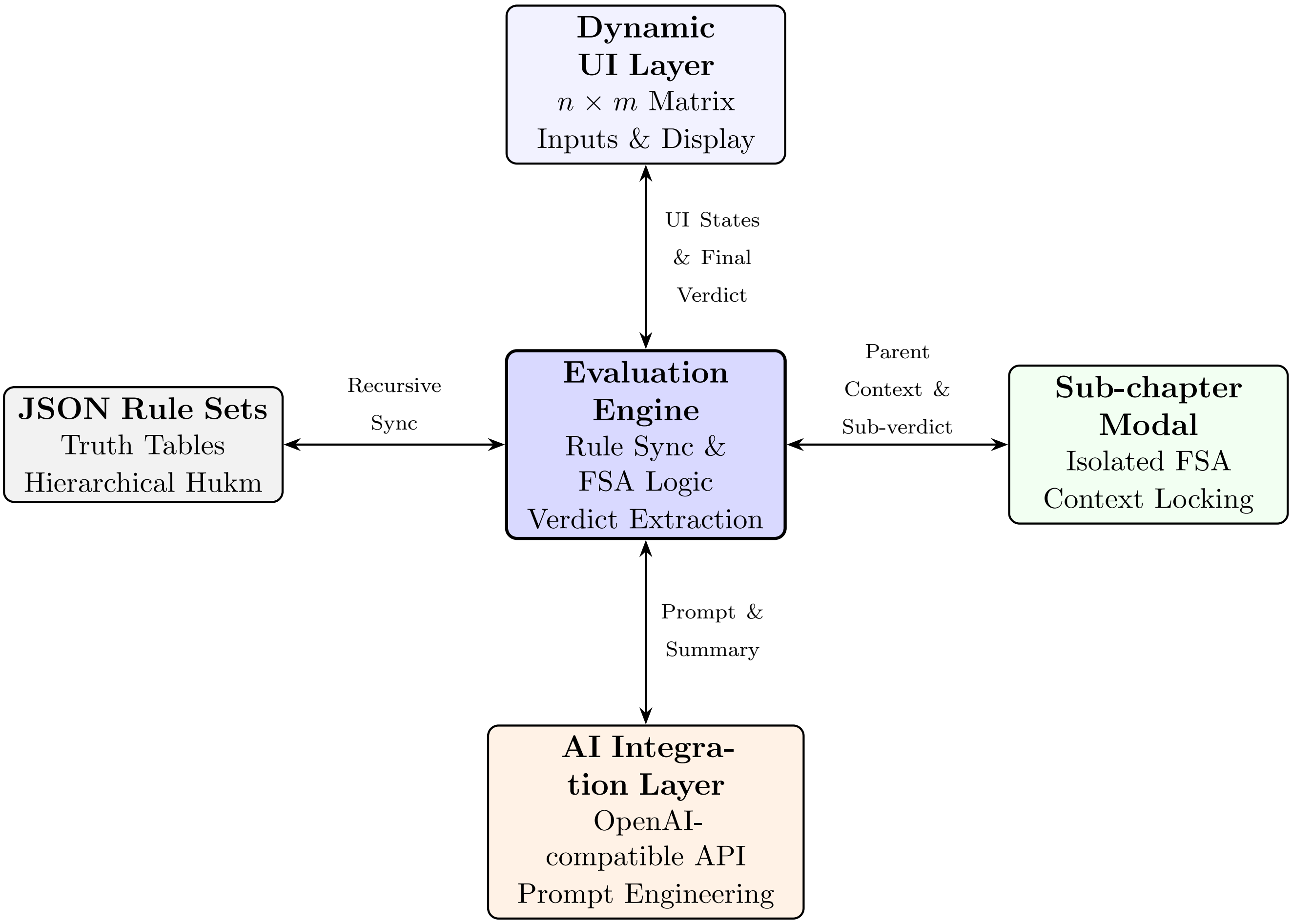

The architecture of the proposed system follows a modular Hub-and-Spoke model. The core Algorithmic Rule Engine acts as the central hub, orchestrating data flow between the User Interface, the hierarchical JSON rule sets, nested sub-chapter evaluators, and the AI integration layer. Figure~\ref{fig:architecture} illustrates the bidirectional data transfer between these components.

\subsection{Dynamic UI Layer}
Unlike static Fiqh applications, the proposed system features a dynamic grid generator. The user defines two parameters:
\begin{itemize}
    \item $m$: The number of horizontal ``Action'' buttons (e.g., Prayed, Defecated).
    \item $n$: The number of vertical ``Event'' buttons (e.g., Time entered, Water lost).
\end{itemize}

Upon generation, the system creates an $n \times m$ matrix of parameter cards. Each card can be toggled between two types:
\begin{enumerate}
    \item \textbf{Simple (\textit{Mufrad}):} A binary toggle representing a direct condition (1 = Present, 0 = Absent).
    \item \textbf{Sub-chapter (\textit{Bab Far'i}):} A nested rule set that opens an isolated Finite State Automaton (FSA) modal for internal evaluation.
\end{enumerate}

\subsection{Rule Representation and Synchronization}
Rules are structured in hierarchical JSON format. The Main Chapter contains variables ($V = \{v_1,\allowbreak v_2,\allowbreak \dots,\allowbreak v_k\}$), a ruling (\textit{Hukm}), and a Truth Table mapping binary strings to specific outcomes and comments. 

A recursive synchronization algorithm (\texttt{findVarInRules}) is implemented to ensure UI consistency. When a user types a parameter name, the algorithm searches the Main JSON and all its nested sub-chapters. If a match is found, the UI automatically locks the parameter type and imports the exact positive/negative Arabic terminology from the JSON into the UI toggles.

\section{Methodology and Algorithmic Evaluation}

\subsection{Bidirectional Communication}
When a sub-chapter is executed, the system packages the current state of all Main UI parameters into a \texttt{parentContext} object. This object is passed to the sub-chapter via the HTML \texttt{<iframe>} \texttt{postMessage} API.

The sub-chapter parses this context and locks any variables that match the parent's state (marked as \textit{Inherited}). Once the user evaluates the sub-chapter, the final binary verdict (1 or 0) and its associated juristic comment are sent back to the Main UI, updating the main state.

\subsection{Extraction of Rulings (FSA Logic)}
The system formalizes juristic extraction using three primary sets derived from the Truth Table:
\begin{align}
    M^+ &= \{\text{states where } Hukm = 1\} \\
    M^- &= \{\text{states where } Hukm = 0\} \\
    S &= \text{Intersection of } M^+ \text{ (Conditions)}
\end{align}

During the initialization phase, exactly one parameter (variable $b$) within the main chapter is designated as the ``Main Parameter''. The overall Fiqh verdict for the chapter is not evaluated continuously, but is strictly drawn when this specific parameter $b$ is toggled to its positive state ($b=1$) during the course of usage. When $b$ is triggered, the system maps the current UI state to a binary string. It checks this string against the $M^+$ array. If a match is found, the ruling is valid; otherwise, it is invalid. The corresponding comment from the Truth Table is retrieved, and the main parameter $b$ is reset to 0.

\section{LLM Integration for Analytical Summarization}
To enhance user comprehension, the system integrates an OpenAI-compatible API endpoint (tested with BlueSminds GLM-4.6). 

Upon triggering the LLM module, the system dynamically constructs a prompt containing the entire \texttt{history} array---a sequential log of all events, actions, and FSA verdicts that occurred during the session. This structured prompt is sent to the LLM via a JavaScript \texttt{fetch} POST request. The LLM acts as a virtual jurist (\textit{Mufti}), analyzing the deterministic state transitions and generating a logical, simplified explanation of \textit{why} the rulings fell as they did, which is then rendered in the UI output box.

\section{Safe Principles of Design of Fikh-Finite State Automata (FFSA)}

The construction and execution of an FFSA within the proposed framework are governed by six foundational design principles. These principles should be understood as sufficient conditions for adherence to the basic safety laws and epistemological boundaries of \textit{Usul al-Fiqh}. By strictly enforcing these constraints, the computational model ensures that its structural inferences do not violate the integrity of Islamic jurisprudential reasoning, while simultaneously maximizing computational efficiency.

\begin{enumerate}
    \item \textbf{Hierarchical Chapter Decomposition:} An FFSA is structurally divided into exactly one Main Chapter (representing the primary ruling, $H$) and an arbitrary number of nested Sub-chapters (\textit{Bab Far'i}). This modular design prevents combinatorial explosion by allowing complex chapters to utilize the binary outputs of atomic sub-chapters as single inputs.
    
    \item \textbf{External Rule File Integration:} The FSA transition logic for both the main chapter and sub-chapters is not hard-coded. It is dynamically loaded from externally saved JSON rule files. These files contain the truth tables and minimal operational rules ($M^+$, $M^-$, $S$, $I$, $W$) generated a priori by the computational \textit{al-sabr wa al-Taqsim} algorithm.
    
    \item \textbf{Main Parameter Initialization:} During the initialization phase of the User Interface, exactly one parameter (variable $b$) within the main chapter is designated as the ``Main Parameter''. The overall Fiqh verdict for the chapter is not evaluated continuously, but is strictly drawn---and the state formally assessed---only when this specific parameter $b$ is toggled to its positive state ($b = 1$) during the course of usage, mimicking the actual execution of the action in question (e.g., the act of praying).
    
    \item \textbf{Sequential Action and Event Modeling:} The FFSA is designed to process temporal sequences of events and actions ($\sigma = \langle x_1, x_2, \dots, x_p \rangle$). The automaton maintains a state history, transitioning deterministically from one state to another as environmental events occur or actions are executed, capturing the dynamic nature of a \textit{Mukallaf's} (legal agent) state over time.
    
    \item \textbf{FSA Supremacy over LLM Integration:} In the system architecture, the deterministic FSA output strictly controls the Large Language Model (LLM) engine, not vice versa. The LLM does not participate in state transitions or ruling calculations. Instead, the compiled sequence of FSA states and verdicts (the history variable) is passed to the LLM solely as a contextual prompt. The LLM acts only as an analytical interpreter, translating the deterministic computational history into a natural language juristic explanation.
    
    \item \textbf{Grammar-Guided LLM Constrained Decoding:} To guarantee that the LLM's natural language generation does not invent juristic facts (hallucination), the LLM output must be strictly controlled via a Constrained Decoding mechanism. The system dynamically constructs formal grammars (such as Context-Free Grammars or Regular Grammars) on the fly from the FFSA execution history ($\Sigma_h$). These grammars mathematically bind the LLM's token generation, ensuring that any terminal legal token generated (representing rulings, conditions, causes, or preventers) is an exact string match to an element actively present in the FSA's deterministic history.
\end{enumerate}

\section{Design Principles of FFSA Guarantee Absolute Absence of Juristic LLM Hallucination}

A primary challenge in integrating Large Language Models (LLMs) with legal or juristic systems is the risk of ``hallucination''---where the probabilistic model generates factually incorrect rulings or invents non-existent legal causes \cite{ji2023}. The proposed FSA-supremacy architecture, augmented with strict Constrained Decoding, mathematically neutralizes this threat. We formalize this by defining the bounds of juristic reality, the grammar governing token generation, and two specific types of hallucination, proving that the architecture prevents both.

\subsection{Formal Definitions and Constrained Grammar}

Let $P$ be the strict Horn-clause program for a closed Fiqh chapter, and let $A = (Q, \Sigma, \delta, q_0, F)$ be its corresponding deterministic FSA. Let the execution history be $\Sigma_h = \langle (x_1, q_1, H_1), (x_2, q_2, H_2), \dots, (x_p, q_p, H_p) \rangle$, where $H_k$ is the deterministic verdict at step $k$. 

Let the LLM be a probabilistic text generation function $L$. At each generation step $t$, the LLM computes a probability distribution over a vocabulary of tokens $V_{tok}$, denoted $P(y_t \mid y_{<t}, \text{Prompt}(\Sigma_h))$. 

To enforce strict grounding, we introduce a formal Context-Free Grammar (or Regular Grammar) $\mathbb{G}_{\Sigma_h}$ dynamically constructed from the execution history $\Sigma_h$ to guide constrained decoding \cite{willard2023}. The grammar $\mathbb{G}_{\Sigma_h}$ defines production rules such that any terminal legal token $\tau$ representing a juristic fact must be an exact string match to an element in the set of active facts in $\Sigma_h$. 

\begin{definition}[Juristic Facts]
Let $\mathcal{J}(\Sigma_h)$ be the finite set of \textit{Juristic Facts} derived from the execution history. This set strictly comprises:
\begin{enumerate}
    \item \textbf{Rulings:} The boolean verdict $H_k$ at any step $k$.
    \item \textbf{Ontological States:} The active binary valuations of the legally operative concepts (OCT) $v_i = b$ at step $k$.
    \item \textbf{Legal Mechanics:} The specific structural components extracted by the \textit{al-\d{S}abr wa al-Taqsim} algorithm active at step $k$, namely: Conditions (\textit{Shurut}, $S$), Causes (\textit{'Ilal}, $I_i$), and Preventers (\textit{Mawani'}, $W_j$).
\end{enumerate}
Linguistic connectors, transitional phrases, and interpretive text not asserting a legal truth value are defined as \textit{Linguistic Artifacts} and are not considered Juristic Facts.
\end{definition}

\begin{definition}[Decisional Hallucination]
A Decisional Hallucination occurs if the LLM output asserts a final ruling $H_{LLM} \in \{0, 1\}$ such that $H_{LLM} \neq H_p$, where $H_p$ is the true FSA verdict at the final step $p$.
\end{definition}

\begin{definition}[Juristic Factual Hallucination]
A Juristic Factual Hallucination occurs if the LLM output contains a juristic proposition $\phi$ asserting a Juristic Fact (e.g., the presence of a condition $S$, a cause $I$, a preventer $W$, or an ontological state $v_i = b$) such that $\phi \notin \mathcal{J}(\Sigma_h)$ (i.e., it is not logically derivable from the execution history or the underlying rule program $P$).
\end{definition}

\subsection{Formal Definition of the Grammar Generation Mapping}
Theorem 1 below relies on the dynamic construction of a formal grammar $G_{\Sigma_h}$ from the FSA execution history $\Sigma_h$. To ensure strict mathematical grounding and deterministic LLM behavior, we define the mapping function $f: \Sigma_h \rightarrow G_{\Sigma_h}$ algorithmically. The function operates in four distinct steps:

\textbf{Step 1: Extraction of Juristic Facts ($J(\Sigma_h)$)}\\
The function $f$ first filters the execution history $\Sigma_h$ to extract the active Juristic Facts (as defined in Definition 1). Let $J(\Sigma_h) = \{ \tau_1, \tau_2, \dots, \tau_k \}$ be the finite set of string literals representing the active ruling, ontological states, and legal mechanics at the final step $p$. Any legal concept not active in $\Sigma_h$ is excluded from this set.

\textbf{Step 2: Static Scaffolding Definition}\\
The system defines a static set of non-terminal symbols $N = \{ S, \text{Ruling}, \text{Illah}, \text{Shart}, \dots \}$ and a static structural template (the start symbol $S$). This scaffolding represents the linguistic structure of a juristic summary but contains no terminal legal vocabulary.

\textbf{Step 3: Dynamic Terminal Injection}\\
The core of the mapping $f$ binds the FSA state to the LLM output space. For every non-terminal $N_i \in N$ representing a specific legal category, $f$ constructs production rules of the form $N_i \rightarrow \tau_j$, where $\tau_j \in J(\Sigma_h)$ and $\tau_j$ belongs to the category of $N_i$. If a fact is not present in $J(\Sigma_h)$, it is not injected as a terminal, rendering its generation probability exactly 0.

\textbf{Step 4: Unambiguity and Determinism Guarantee}\\
To ensure the Constrained Decoding engine can parse the grammar without conflicts, $f$ is restricted to generating a \textit{Right-Linear Grammar} (a strict subset of Context-Free Grammars)\cite{hopcroft2006}. Every production rule is of the form $A \rightarrow aB$ or $A \rightarrow a$, where $A, B \in N$ and $a$ is a terminal. Because right-linear grammars are equivalent to Deterministic Finite Automata (DFA), the generated grammar $G_{\Sigma_h}$ is strictly unambiguous. It can be parsed deterministically in linear time, guaranteeing that the LLM decoding process will never encounter parse conflicts or dead ends.

\subsubsection*{Concrete Example: Grammar Generation for a Tayammum Ruling}
To illustrate the mapping $f$, consider a user querying the system about the validity of their Tayammum (dry ablution). Suppose the FSA processes the inputs and reaches a final state $q_p$ where the verdict is valid due to a lack of water and the presence of dust. 

The execution history $\Sigma_h$ yields the following active Juristic Facts:
\begin{itemize}
    \item \textbf{Ruling:} $H_p = 1$ (mapped to string literal \texttt{"Valid"})
    \item \textbf{Cause (Illah):} \texttt{"absence of water"}
    \item \textbf{Condition (Shart):} \texttt{"presence of dust"}
\end{itemize}

The mapping $f(\Sigma_h)$ generates the following unambiguous right-linear grammar $G_{\Sigma_h}$, where the start symbol is $S$:

\begin{align*}
S &\rightarrow \text{"Your Tayammum is"} \quad \text{Ruling} \\
\text{Ruling} &\rightarrow \text{"Valid"} \quad \text{Connector1} \\
\text{Connector1} &\rightarrow \text{"because of the"} \quad \text{Illah} \\
\text{Illah} &\rightarrow \text{"absence of water"} \quad \text{Connector2} \\
\text{Connector2} &\rightarrow \text{"and the"} \quad \text{Shart} \\
\text{Shart} &\rightarrow \text{"presence of dust"} \quad \text{End} \\
\text{End} &\rightarrow \text{"."}
\end{align*}

During LLM text generation, the Constrained Decoding engine applies a binary mask $M_t$ at each step based strictly on $G_{\Sigma_h}$. If the LLM's underlying neural network proposes the token \texttt{"Invalid"} (due to training bias) at the \text{Ruling} state, $M_t$ sets its probability to 0, forcing the selection of \texttt{"Valid"}. Similarly, if the LLM proposes a hallucinated cause like \texttt{"eating camel meat"} at the \text{Illah} state, the mask sets it to 0, as the only valid transition is \texttt{"absence of water"}. 

By formalizing $f$ as a right-linear grammar generator, the system guarantees that the LLM is mathematically bound to traverse only the paths justified by $\Sigma_h$, reducing the model to a pure linguistic transformer of deterministic facts.

\subsection{Theorem and Proof}

We now state and prove the theorem guaranteeing the absolute absence of both types of hallucination.

\begin{theorem}[Strict Juristic Grounding via Constrained Decoding]
Under the FFSA design principles, utilizing a Constrained Decoding mechanism governed by grammar $\mathbb{G}_{\Sigma_h}$, the architecture guarantees the absolute absence of Decisional Hallucination and Juristic Factual Hallucination. The operational juristic truth of the system is mathematically isolated from probabilistic LLM errors.
\end{theorem}

\begin{proof}
The system is modeled as $S = (A, L, \mathbb{G}_{\Sigma_h})$.

\textbf{1. Immunity to Decisional Hallucination:}
The FSA state transition is defined as $q_p = \delta(q_{p-1}, x_p)$ and the final system verdict is strictly determined by checking if $q_p \in F$. By Design Principle 5 (FSA Supremacy), the LLM $L$ does not participate in the transition function $\delta$, the accepting states $F$, or the input variables $x_p$. Because there is no feedback loop from $L$ to $A$, the LLM possesses zero write-access to the system state. The system's authoritative output is strictly governed by $A$, yielding $H_p$. Therefore, even if the LLM generates text asserting a false ruling, the system's operational state and final verdict remain strictly $H_p$. Decisional hallucination is architecturally impossible and cannot affect the user's received verdict.

\textbf{2. Absolute Absence of Juristic Factual Hallucination:}
By the definition of Constrained Decoding, at each generation step $t$, the LLM computes a probability distribution over the vocabulary $P(y_t \mid y_{<t}, \text{Prompt}(\Sigma_h))$. The decoding algorithm applies a binary mask $M_t$ derived from the grammar $\mathbb{G}_{\Sigma_h}$:
 $$ P_{\text{constrained}}(y_t) = P(y_t \mid y_{<t}, \text{Prompt}(\Sigma_h)) \cdot M_t(\mathbb{G}_{\Sigma_h}, y_{<t}) $$ where $M_t$ sets the probability of any token that would violate the production rules of $\mathbb{G}_{\Sigma_h}$ to exactly $0$.

Because grammar $\mathbb{G}_{\Sigma_h}$ is constructed such that terminal nodes representing Juristic Facts are strictly drawn from the set $\mathcal{J}(\Sigma_h)$, the probability of the LLM generating a terminal legal token sequence $\tau$ representing a Juristic Fact $\phi \notin \mathcal{J}(\Sigma_h)$ is exactly $0$. 

Consequently, the LLM is physically incapable of outputting a false condition, an invented cause (\textit{'illah}), a non-existent preventer (\textit{mani'}), or an incorrect boolean state. Every Juristic Fact $\phi$ present in the generated text $T$ is strictly entailed by $\Sigma_h$. Thus, Juristic Factual Hallucination is mathematically impossible.
\end{proof}

\begin{remark}[Containment of Linguistic Artifacts]
While the architecture mathematically guarantees that no false Juristic Facts can be generated, the LLM retains the probabilistic capacity to generate ``Linguistic Artifacts'' (e.g., conversational connective tissue or historical quotes not present in $\Sigma_h$). However, because these artifacts do not constitute Juristic Facts as defined in Definition 1, and because they are never fed back into the FSA, they cannot alter the legal state space $Q = \{0, 1\}^n$ or the final ruling $H_p$. The ground truth of the system is strictly anchored to the compiled Horn-clause program $P$, entirely bypassing the LLM's probabilistic inference for legal determinations.
\end{remark}

\subsection{Formal Containment of Semantic Inversion}
While Theorem 1 proves that the Constrained Decoding mechanism prevents the generation of false Juristic Facts, Remark 1 permits the LLM to generate unconstrained ``Linguistic Artifacts'' (connective tissue). A theoretical vulnerability remains: \textit{Semantic Inversion}. Semantic inversion occurs when the LLM uses unconstrained connectors to syntactically surround a constrained legal token with negating or contradictory context. 

For example, if the FSA determines the verdict is ``Valid'', an unconstrained LLM might generate: 
\begin{itemize}
    \item \textit{Inverted Output:} ``Your Tayammum is valid, \textbf{however}, this is a universally rejected view.''
\end{itemize}
While the token ``valid'' is syntactically permitted, the adversative connector ``however'' semantically inverts the pragmatic meaning of the sentence, bypassing the syntactic mask.

To mathematically eliminate this threat, the grammar generation mapping $f$ is further constrained by a \textbf{Bounded Neutral Lexicon} and \textbf{Valence Locking}.

\textbf{1. Bounded Neutral Lexicon:}\\
The non-terminal transition states in the right-linear grammar $G_{\Sigma_h}$ (such as \text{Connector1}, \text{Connector2}) are not left unconstrained. They are restricted to drawing terminals from a finite, pre-defined lexicon of semantically neutral connectors, denoted $\mathcal{L}_{neutral}$. This set includes phrases such as ``is'', ``because of'', ``and'', ``due to'', and ``the''. Negating, contrasting, or adversative connectors (e.g., ``however'', ``but'', ``despite'', ``not'', ``although'') are strictly excluded from the production rules. If a token is not in $\mathcal{L}_{neutral}$, the Constrained Decoding mask $M_t$ sets its generation probability to 0.

\textbf{2. Valence Locking:}\\
Because the only permissible paths between Juristic Facts in $G_{\Sigma_h}$ are neutral connectors from $\mathcal{L}_{neutral}$, the semantic polarity (valence) of the ruling $H_p$ is mathematically locked. The LLM is structurally incapable of constructing a sentence that flips the truth value of the embedded legal fact. 

Using the previous example, if the FSA verdict is ``Valid'' and the active cause is ``absence of water'', the constrained grammar forces the LLM to generate a valence-locked output, such as:
\begin{itemize}
    \item \textit{Valence-Locked Output:} ``Your Tayammum is valid \textbf{because of} the absence of water.''
\end{itemize}

By restricting the transition space to a neutral valence, the architecture guarantees that the generated text $T$ is semantically faithful to the deterministic verdict $H_p$. Therefore, the probability of the LLM generating a sentence that semantically contradicts the FSA state is exactly 0. The system guarantees not only syntactic grounding (absolute absence of false facts) but also semantic grounding (absolute absence of false meanings), fully neutralizing the threat of LLM hallucination in juristic summarization.

\section{Conclusion}
Al-Ghasali successfully demonstrates that complex Islamic jurisprudential rulings can be dynamically modeled, evaluated, and explained using modern web technologies. By proving the finite-state collapse of closed Horn chapters, we established that restricting Fiqh formalization to finite propositional theories guarantees decidability and constant-time (O(1)) verdict extraction, avoiding the computational burdens of recursive Datalog or Turing-complete systems. The system compiles the outputs of the al-sabr wa al-Taqsim algorithm into deterministic FSA state tables, achieving highly efficient, real-time juristic simulation. Furthermore, by mathematically formalizing temporal sequences such as \textit{Tartib} (juristic order) as static ontological variables and proving the decidability of hierarchical sub-chapter composition, the framework ensures that complex, multi-layered, and path-dependent rulings remain strictly bounded by O(1) extraction guarantees.

A major outcome of this research is the formulation of sufficient design safety conditions governing the construction and execution of Fiqh-Finite State Automata (FFSA). These principles ensure that the structural inferences of the computational model strictly adhere to the basic safety laws and epistemological boundaries of \textit{Usul al-Fiqh}. By enforcing FSA supremacy over the LLM and implementing Grammar-Guided Constrained Decoding---formalized as a right-linear grammar generator augmented by a Bounded Neutral Lexicon and Valence Locking---the architecture mathematically guarantees the absolute absence of Decisional Hallucination, Juristic Factual Hallucination, and Semantic Inversion. Al-Ghasali safely reduces the LLM to a pure linguistic transformer of the deterministic FSA output, entirely isolating the operational juristic truth of the system from probabilistic LLM errors. This provides a robust, mathematically proven tool for both educational and research purposes in computational Fiqh. Future work could involve wrapping the application in a standalone desktop environment and expanding the system's scope to handle bounded Datalog-style recursive relations, such as genealogical kinship, while preserving the framework's core efficiency and strict semantic grounding.


\begin{thebibliography}{13}

\bibitem{abdelwahab2026}
E. Abdelwahab, ``Computational Extraction of Legal Causes via al-sabr wa al-Taqsim: A Set-Theoretic Formalization for Closed Fiqh Chapters,'' \emph{Journal Academica}, vol. 15, no. 1, pp. 1--16, 2026.

\bibitem{lloyd1987}
J. W. Lloyd, \emph{Foundations of Logic Programming}, 2nd ed. Berlin, Germany: Springer-Verlag, 1987.

\bibitem{hopcroft2006}
J. E. Hopcroft, R. Motwani, and J. D. Ullman, \emph{Introduction to Automata Theory, Languages, and Computation}, 3rd ed. Boston, MA, USA: Addison-Wesley, 2006.

\bibitem{sipser2013}
M. Sipser, \emph{Introduction to the Theory of Computation}, 3rd ed. Boston, MA, USA: Cengage Learning, 2013.

\bibitem{ceri1990}
S. Ceri, G. Gottlob, and L. Tanca, \emph{Logic Programming and Databases}. Berlin, Germany: Springer-Verlag, 1990.

\bibitem{ji2023}
Z. Ji, N. Lee, R. Frieske, T. Yu, D. Su, Y. Xu, E. Ishii, Y. J. Bang, A. Madotto, and P. Fung, ``Survey of Hallucination in Natural Language Generation,'' \emph{ACM Computing Surveys}, vol. 55, no. 12, pp. 1--38, 2023.

\bibitem{willard2023}
B. T. Willard and A. Vetta, ``Efficient Guided Generation for Large Language Models,'' \emph{arXiv preprint arXiv:2307.09702}, 2023.

\bibitem{alshatibi1997}
I. M. Al-Shatibi, \emph{Al-Muwafaqat fi Usul al-Shari'a}, vol. 1. Riyadh, Saudi Arabia: Dar Ibn Affan, 1997.

\bibitem{kamali2003}
M. H. Kamali, \emph{Principles of Islamic Jurisprudence}, 3rd rev. ed. Cambridge, UK: Islamic Texts Society, 2003.

\bibitem{alamidi2000}
S. Al-Amidi, \emph{Al-Ihkam fi Usul al-Ahkam}, vols. 1--4. Beirut, Lebanon: Dar al-Kutub al-Ilmiyya, 2000.

\bibitem{alshawkani1999}
M. Al-Shawkani, \emph{Irshad al-Fuhul ila Tahqiq al-Haqq min Ilm al-Usul}. Beirut, Lebanon: Dar al-Kutub al-Ilmiyyah, 1999.

\bibitem{alnawawi2002}
M. Al-Nawawi, \emph{Al-Majmu' Sharh al-Muhadhdhab}, vol. 2. Beirut, Lebanon: Dar al-Kutub al-Ilmiyyah, 2002.

\bibitem{almarghinani2004}
B. Al-Marghinani, \emph{Al-Hidayah: The Guidance}, translated by C. Hamilton. Leiden, Netherlands: Brill, 2004.

\end{thebibliography}
\end{document}